\documentclass[runningheads]{llncs}
\usepackage{graphicx}
\begin{document}
\title{Evaluating the Predictive Features of Person-Centric Knowledge Graph Embeddings: Unfolding Ablation Studies}
\titlerunning{Evaluating the Predictive Features of PKG Embeddings}
%
\author{Christos Theodoropoulos\inst{1}
        \orcidID{0000-0002-2810-7817} 
        \and
        Natasha Mulligan\inst{2}
        \orcidID{0009-0006-5631-0325} 
        \and
        Joao Bettencourt-Silva\inst{2}
        \orcidID{0000-0002-4562-0911}
        }

\authorrunning{Christos Theodoropoulos et al.}
%
\institute{KU Leuven, Leuven, Belgium \\
           \email{\{christos.theodoropoulos\}@kuleuven.be}
           \and
           IBM Research Europe, Dublin, Ireland \\
           \email{\{jbettencourt, natasha.mulligan\}@ie.ibm.com}
}
\maketitle              
\begin{abstract}

Developing novel predictive models with complex biomedical
information is challenging due to various idiosyncrasies related to heterogeneity, standardization or sparseness of the data. We previously introduced a person-centric ontology to organize information about individual patients, and a representation learning framework to extract person-centric knowledge graphs (PKGs) and to train Graph Neural Networks (GNNs). In this paper, we propose a systematic approach to examine the results of GNN models trained with both structured and unstructured information from the MIMIC-III dataset. Through ablation studies on different clinical, demographic, and social data, we show the robustness of this approach in identifying predictive features in PKGs for the task of readmission prediction.

\end{abstract}
\section{Introduction and Background}
Arguably the most sought-after predictive models in healthcare are those aimed at predicting readmissions. Avoiding subsequent, unplanned, hospitalizations has been a longstanding multidisciplinary effort involving clinicians, informaticians, researchers, statisticians as well as policymakers. Healthcare systems have introduced tracking metrics and incentives, with particular emphasis on conditions that are more prone to readmissions such as cardiovascular events or pneumonia \cite{khan2021trendsfull}. Indeed nearly 1 in 4 heart failure (HF) patients are readmitted within 30 days, and approximately half of HF patients are readmitted within 6 months. Globally, readmissions are on the increase and reducing 30-day readmissions has now been a longstanding target to improve the quality of care and reduce expenditure \cite{khan2021trendsfull,lawson2021trends,mcneill2020impact}. Identifying the most relevant predictors has been reportedly challenging, especially for conditions such as HF \cite{sheetrit2023predictingfull,10.1145/3587259.3627545full}. Recent efforts have sought to bring additional variables, parameters, and modalities to better understand the key drivers of readmissions. For example, it has been shown that notes written by physicians may have better capabilities for readmission prediction when compared with other modalities \cite{sheetrit2023predictingfull}. Other efforts have focused on creating a richer picture of patients by including sociodemographic features, such as social determinants of health (SDoH), which are often embedded in clinical notes \cite{bettencourt2020discoveringfull}. Finally, a recent approach focused on using neural networks and process mining for detecting important predictors based on time information associated with hospital events, severity scores, and demographics \cite{pishgar2022predictionfull}. 
\par
Graph Neural Networks (GNNs) \cite{10.1145/3587259.3627545full,wu2020comprehensivefull} have emerged as powerful tools for learning representations from graph-structured data from various domains. By leveraging the inherent structure and connectivity of graphs, GNNs offer a versatile approach for capturing complex relationships. However, methodological frameworks to continuously evaluate model results with respect to a given prediction task are needed, especially in healthcare domains. This paper proposes a method to systematically evaluate expert-driven hypotheses about predictors of interest and their impact on GNN model results. Our approach centers on learning representations for person-centric graphs using GNNs, and, by encoding the intricate interplay between patients' clinical attributes,
demographics, and social aspects within the graph structure, we have developed GNN-based predictive models capable of identifying patients at high risk of readmission.

\section{Methods}

We previously developed an ontology that defines person-centric knowledge graphs (PKGs) and their social context, and studied how different ontology-driven graph representations and heterogeneity impact models’ performance \cite{10.1145/3587259.3627545full}. In this paper, we introduce a repeatable method and approach to continuously identify predictive features of interest based on ablation studies (Figure 1A). The approach requires as input: a task (e.g., readmission prediction), a dataset (e.g., hospital health records), and a person-centric graph schema (e.g., ontology defining the available data points and how they relate to a central node of interest). Before training a GNN, the following pre-processing steps are required to create PKGs: (1) data must first be selected from the input dataset, (2) an assessment of the data and its quality (e.g. missing data) is performed, (3) data can
further be enriched by mining unstructured text notes or introducing other modalities, (4) the data is then sampled and (5) a summary of the dataset is created so that knowledge graphs can be extracted. Our framework for GNN model training requires individual graphs to be created for each patient hospital admission. Once the model is trained, an extensive ablation study is conducted to unearth the most predictive features encoded within the graph representation. Finally, an analysis and interpretation of model results is carried out to dissect the contributions of different features, such as demographic variables and social determinants of health, in the predictive performance of the model.

\subsection{Pre-processing and Graph Neural Network Model Training}

The selected task was predicting Intensive Care Unit (ICU) readmissions using the MIMIC-III dataset \cite{johnson2016mimic}. In this paper, we rely on a representation learning framework previously developed \cite{10.1145/3587259.3627545full} and the Health \& Social Patient-centric Ontology (HSPO) to define how PKGs are created. Readmissions are defined as any subsequent admissions within 30 days of the first ICU visit in MIMIC-III. Data was extracted such that PKGs contain information about demographic, clinical, and social context, each represented by distinct graph nodes. The demographic nodes encapsulate attributes such as age group, gender, religion, marital status, and race. For the clinical view, nodes represent diseases, received medication, and undergoing procedures. Additionally, to incorporate social
aspects, information regarding employment status, household composition, and housing conditions is extracted from clinical notes using UMLS codes sampled via the MetaMap
annotator\footnote{MetaMap. 2016. \\ \url{https://www.nlm.nih.gov/research/umls/implementation\_resources/metamap.html}} \cite{aronson2001effective} and represented as nodes in the graphs.

\begin{figure}[!t]
  \centering
  \includegraphics[width=\textwidth]{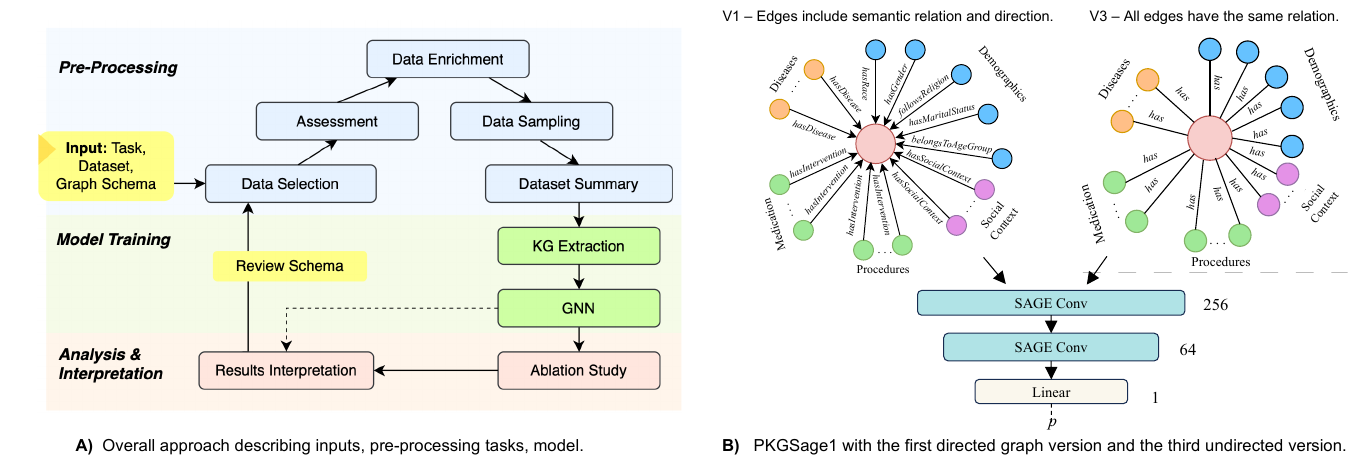}
   \caption{Overall approach (A), Knowledge Graphs and GNN (B).}
\end{figure}

\par
We utilize the best-performing model PKGSage1 \cite{10.1145/3587259.3627545full} that leverages the Sage Graph Convolution Network architecture \cite{hamilton2017inductive} (Figure 1B). Determining the most effective graph structure is a challenging task and was previously addressed in \cite{10.1145/3587259.3627545full}. We conducted experiments using the directed first graph version and the undirected third version (Figure 1B), both of which demonstrate the best performance.

\subsection{Ablation Study}

The ablation study aims at comprehensively assessing the predictive ability of different features and facets (i.e. group of features) within the person-centric graph embeddings. The objective is to systematically evaluate the impact of excluding different information on the model’s predictive performance. Initially, each facet is removed, and the resulting performance degradation is assessed. To obtain further insights, different combinations of features (e.g. all medications, procedures) are also excluded. Given the complex interdependencies between clinical information, we explore the robustness of the models by conducting experiments at a second level, where two facets of the clinical view are entirely excluded, followed by the exclusion of all clinical facets (diseases, medication, and procedures). Furthermore, considering the ICU readmission prediction task, further experiments are conducted with a focus on excluding the most well-known and frequent conditions in U.S. hospitals for all-cause 30-day readmissions\footnote{Statistical Brief 153. HCUP. May 2016. US AHRQ. \url{https://hcup-us.ahrq.gov/reports/statbriefs/sb153.jsp}}.

\section{Results and Discussion}

The results of the ablation study (Figure 2A, B) shed light on the importance of demographic information, particularly in the context of readmission prediction models. We observed that the removal of socio-demographic variables impacts model
performance, particularly in terms of accuracy. Among these, marital status and race emerged as the two strongest predictors within our dataset. It is noteworthy that marital status, in addition to its demographic significance, also carries a social dimension pertaining to the patient's support network and living arrangements. Contrary to initial
expectations, age did not emerge as the strongest predictor in our analysis. However, a deeper examination of the dataset revealed that a significant proportion of patients are elderly, with 56.24\% being over 70. This underscores the unique demographic composition of the dataset and emphasizes the need for a nuanced analysis.
\par 
Despite the large number of missing social information, our study demonstrated that excluding social aspects detrimentally affects model performance. The exclusion of household information (77.9\% missing data) had the most performance impact. This finding underscores the critical role of social determinants in predictive models and the
importance of robust data collection strategies. While the systematic acquisition of social data poses challenges, such as dealing with unstructured data \cite{theodoropoulos-etal-2021-imposingfull}, extracting knowledge from biomedical text \cite{10.1007/978-3-031-39965-749} and defining a standardized protocol, it is imperative for enhancing a data-driven understanding of readmission prediction or other tasks.
\par 
Regarding clinical predictors (Figure 2B), we find that the pure clinical view of patients emerges as a key driver, with diseases and medications standing out as the most influential factors. One notable observation is the inherent inter-correlations among various aspects of the clinical view (e.g., patients diagnosed with specific diseases will
receive disease-specific medications and procedures). While excluding the most common diagnoses associated with high readmission rates has a discernible effect on model performance, the exclusion of the entire disease information leads to an additional drop in performance metrics, up to 4.13\% percentage decrease in accuracy. This suggests
that additional diagnoses beyond the most known ones may play a substantial role in driving ICU readmissions. These findings underscore the importance of expanding the scope of diagnoses (and ICD code mappings) considered in predictive modeling efforts to capture a more comprehensive view of readmission risk factors. Medication and procedures are also strong predictors (Figure 2B).
\par 
Regarding the exclusion of paired clinical information, such as medication-procedure or disease-medication combinations, we observe a further impact on model efficiency, albeit not as significant as the exclusion of entire facets. This reaffirms the intercorrelation among clinical factors and underscores the need for holistic modeling approaches. Excluding clinical information completely results in significant performance degradation with up to 18.24\% and 8.37\% percentage decline in accuracy and F1 score (i.e., the harmonic mean of precision P and recall R: $F_1 = 2\frac{P*R}{P+R}$) respectively. Hence, this supports the fact that the clinical view of the patient may be the best predictor for the ICU readmission prediction task given the available data.

\begin{figure}[!t]
  \centering
  \includegraphics[width=\textwidth]{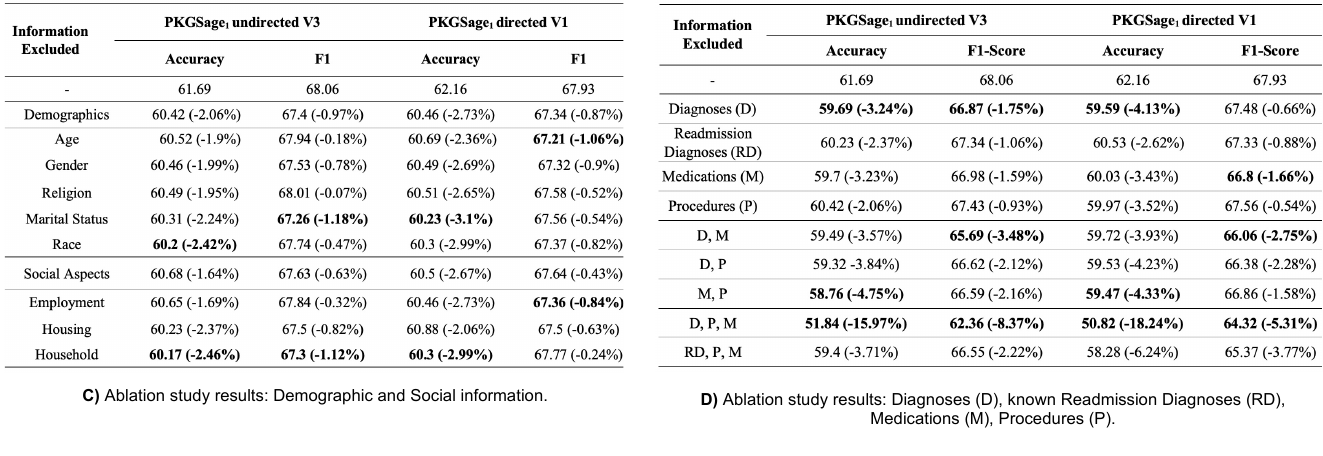}
   \caption{Ablation study results (A, B).}
\end{figure}

\par
Overall, we highlight that the resilience of the models in managing missing data is profound, as the decline in performance is marginal when we exclude one facet of the data. In the worst case, accuracy and F1 score present 4.13\% and 1.75\% percentage decrease respectively. The models exhibit robust performance even when two of the three
clinical facets are not present, with up 4.75\% and 3.48\% percentage deterioration in accuracy and F1 score correspondingly. We emphasize that stability and resilience are crucial characteristics, particularly in healthcare datasets where missing data is unavoidable due to privacy concerns, system limitations, or human errors.

\section{Conclusions}
This paper presents a novel and data-driven approach to identify GNN model predictors through conducting ablation studies within the framework of patient-centric graph prediction tasks. Through an extensive analysis focused on the ICU readmission prediction task, we identify and highlight the most influential predictors, shedding light
on critical factors that impact patient outcomes in clinical settings. One of the key takeaways is the paramount importance of systematic collection of social determinants data, as we detect performance degradation even if this information availability is largely missing in the MIMIC-III dataset. Our framework provides flexible and adaptable
methodological steps that can be applied to different datasets and predictive tasks within the healthcare domain and beyond. These steps allow the continuous examination of new predictors, which can be domain expert-driven, or as future work, using generative models. Future work will also extend the application of our framework to diverse healthcare scenarios and hypotheses. By leveraging the insights from the ablation studies, we aim to support the advancement of robust predictive models that can be interpretable and actionable, ultimately contributing to improved care and service delivery.

%
%
%
\bibliographystyle{splncs04}
\bibliography{mybibliography}
\end{document}